\documentclass[letterpaper]{article} % DO NOT CHANGE THIS
\usepackage{aaai2027}  % DO NOT CHANGE THIS
\usepackage[hyphens]{url}  % DO NOT CHANGE THIS
\usepackage{graphicx} % DO NOT CHANGE THIS
\usepackage{natbib}  % DO NOT CHANGE THIS AND DO NOT ADD ANY OPTIONS TO IT
\usepackage{caption} % DO NOT CHANGE THIS AND DO NOT ADD ANY OPTIONS TO IT
\usepackage{algorithm}
\usepackage{algorithmic}
\usepackage{graphicx}
\usepackage{subcaption}
\usepackage{newfloat}
\usepackage{listings}
\DeclareCaptionStyle{ruled}{labelfont=normalfont,labelsep=colon,strut=off} % DO NOT CHANGE THIS
\floatstyle{ruled}
\newfloat{listing}{tb}{lst}{}
\floatname{listing}{Listing}
\usepackage{xcolor}
\definecolor{linkblue}{RGB}{0,102,204}
\definecolor{neuripspink}{RGB}{214, 92, 132}
\usepackage{amsmath}
\usepackage{amssymb}

\usepackage{booktabs}

\usepackage[hyphens]{url}  % DO NOT CHANGE THIS
\usepackage{graphicx} % DO NOT CHANGE THIS
\usepackage{natbib}  % DO NOT CHANGE THIS AND DO NOT ADD ANY OPTIONS TO IT
\usepackage{caption} % DO NOT CHANGE THIS AND DO NOT ADD ANY OPTIONS TO IT
\usepackage{algorithm}
\usepackage{algorithmic}
\usepackage{graphicx}
\usepackage{subcaption}
\usepackage{newfloat}
\usepackage{listings}
\DeclareCaptionStyle{ruled}{labelfont=normalfont,labelsep=colon,strut=off} % DO NOT CHANGE THIS
\floatstyle{ruled}
\newfloat{listing}{tb}{lst}{}
\floatname{listing}{Listing}

\usepackage{amsmath}
\usepackage{amssymb}

\usepackage{booktabs}

\usepackage[most]{tcolorbox}
\definecolor{cok}{RGB}{27,120,55}
\definecolor{cbad}{RGB}{178,40,40}
\definecolor{cneu}{RGB}{80,80,80}
\newtcolorbox{solbox}[2][]{breakable, enhanced, colback=#2!5, colframe=#2,
  coltitle=white, fonttitle=\bfseries\footnotesize, fontupper=\footnotesize,
  boxrule=0.7pt, arc=2pt, left=5pt, right=5pt, top=3pt, bottom=3pt, title={#1}}
\newcommand{\hlbad}[1]{\colorbox{cbad!14}{#1}}
\newcommand{\hlok}[1]{\colorbox{cok!16}{#1}}

\title{dOPSD: On-Policy Self-Distillation for Diffusion Language Models}
\author{
    Phuong Tuan Dat\textsuperscript{\rm 1},
    Qi Li\textsuperscript{\rm 1},
    Xinchao Wang\textsuperscript{\rm 1}\corresponding
}

\affiliations{
    \textsuperscript{\rm 1}National University of Singapore\\
    \{phuongtuandat, liqi\}@u.nus.edu,
    xinchao@nus.edu.sg \\
\texttt{\textbf{\textcolor{neuripspink}{https://github.com/tuandattt/dOPSD}}}
}

\makeatletter
\def\copyright@on{T}
\def\copyright@text{}
\makeatother
\begin{document}

\maketitle

\begin{abstract}
Diffusion large language models (dLLMs) generate text by iteratively denoising a masked sequence, offering a parallel alternative to autoregressive models, but eliciting strong reasoning through post-training remains difficult: supervised fine-tuning is off-policy and suffers from exposure bias, while reinforcement learning gives only sparse, sequence-level rewards and is hard to apply without tractable sequence likelihoods. On-policy self-distillation (OPSD) offers a promising alternative, using one model as both student and teacher to provide dense, token-level, on-policy supervision, but its effectiveness hinges on giving the teacher privileged information (PI) - typically an instance-specific ground-truth reference unavailable at inference - so the student ends up distilling a weak PI-free consensus policy that yields little improvement on dLLM reasoning. We introduce dOPSD, which instead derives the teacher's privilege directly from the student's own denoising trajectory, evaluating masked positions using later, more-decoded steps of that same trajectory rather than an external label, so the teacher's advantage emerges from the model's own decoding process; on Dream and LLaDA, dOPSD improves both in-domain math reasoning and out-of-domain code generation, outperforming supervised and on-policy baselines.
\end{abstract}


\section{Introduction}

Diffusion large language models (dLLMs) have recently emerged as a competitive, non-autoregressive alternative to standard left-to-right language models~\cite{yu2025discrete, song2025seeddiffusion}. Rather than generating one token at a time, a dLLM begins from a fully masked sequence and produces text through an iterative denoising process: at each step it predicts the clean tokens at the masked positions and commits the most confident ones, progressively filling in the sequence. This paradigm scales to billions of parameters and rivals autoregressive models on general language tasks, while offering parallel decoding and bidirectional context. Yet eliciting strong reasoning from dLLMs through post-training remains challenging. Supervised fine-tuning on reference solutions is off-policy and suffers from exposure bias, while reinforcement learning with verifiable rewards (RLVR) \cite{RLVR, yang2026darediffusionlargelanguage} is costly, supplies only a sparse sequence-level reward, and is itself hard to adapt to diffusion models that lack a tractable sequence likelihood. This motivates a post-training signal that is simultaneously dense, on-policy, and free of external teachers or reward models.

On-policy self-distillation (OPSD) was recently proposed for autoregressive LLMs as exactly such a signal~\cite{zhao2026opsd}. A single model plays two roles that differ only in their conditioning context: a student that sees only the problem, and a teacher that is additionally conditioned on PI, a ground-truth reference solution. The student generates an on-policy rollout, and is trained to match the teacher's dense, per-token distribution along that rollout, needing neither a larger external teacher nor a reward model. The teacher's strength, however, comes entirely from an external, instance-specific label that the student never sees at inference. Recent analysis shows this is more than a practical nuisance: unable to condition on the PI, OPSD ends up optimizing a weak, PI-marginalized ``consensus'' of the per-problem teachers rather than any single strong teacher~\cite{zhu2026manyfaces}. The key, then, is not to discard the privileged teacher, but to source its privilege from something the model itself produces, keeping the dense, on-policy signal while removing the dependence on external reference solutions or chain of thought (CoT)~\cite{chain-of-thought}.

Our central observation is that diffusion decoding supplies such privilege intrinsically. As a dLLM unmasks its sequence, the student's own denoising trajectory becomes a sequence of progressively more informed contexts: a later, more-unmasked snapshot reveals tokens an earlier step had to predict from scratch. A teacher positioned later along the trajectory therefore enjoys a genuine ``peek-ahead'' advantage over the student positioned earlier, about the very positions still to predict. This context is on-policy and self-generated rather than an external label, and is unique to diffusion: an autoregressive model, committing tokens once and left to right, has no comparable ladder of increasingly-informed views of one generation.

We instantiate this idea as \textbf{dOPSD}, an on-policy self-distillation method for diffusion language models whose PI is drawn from the student's own denoising trajectory rather than an external reference solution. The student rolls out a completion while recording its decoding history; at an intermediate, heavily masked step it predicts the still-masked positions, and the same model, acting as the teacher, evaluates those same positions from \emph{later} steps of that trajectory, where more surrounding tokens have already been decoded and the teacher is therefore better informed about what the masked positions should be. We distill the teacher's distribution into the student at the masked positions with a token-level Jensen--Shannon objective. The intermediate trajectory state is simply the completion with the undecoded positions re-masked, so it slots directly into the standard dLLM forward pass with no architectural change. To keep the self-distillation signal trustworthy, we verify each rollout against the final answer and distill only from correct rollouts. The main contributions of this work are summarized as follows:
\begin{itemize}
    \item We show that the original OPSD recipe does not transfer to diffusion language models: its reliance on an external, instance-specific reference solution as PI, which is absent at inference and aggregated into a weak PI-marginalized consensus, fails to yield reliable gains for dLLMs.
    \item We propose \textbf{dOPSD}, a novel post-training method for dLLMs that retains the privileged-teacher principle of OPSD but sources the privilege from the student's own denoising trajectory, making it effective under diffusion decoding and able to outperform supervised, reinforcement-learning, and on-policy self-distillation baselines.
    \item We conduct comprehensive experiments on Dream and LLaDA, showing that dOPSD improves performance both on in-domain mathematical reasoning and on out-of-domain code generation, and we analyze how the teacher's peek-ahead horizon and rollout verification shape the learning signal.
\end{itemize}

\section{Related Work}

\subsection{Diffusion Language Models}
Discrete and masked diffusion models generate sequences by reversing a corruption process that progressively masks tokens, predicting the clean tokens at masked positions instead of decoding strictly left to right~\cite{austin2021d3pm,lou2023sedd,sahoo2024mdlm}. Recent open models such as LLaDA~\cite{nie2025llada} and Dream~\cite{ye2025dream} scale this paradigm to billions of parameters, retaining parallel, confidence-ordered decoding and bidirectional context. Post-training these models for reasoning is an active area: several works adapt RLVR to diffusion language models, where the absence of a tractable sequence log-likelihood has prompted likelihood surrogates, inpainting-based guidance, and sequence-level formulations~\cite{zhao2025d1,zhao2025igpo,ou2025principled}. These methods inherit RLVR's sparse, sequence-level reward; dOPSD instead provides a dense, token-level signal on the model's own decoding trajectory.

\subsection{Knowledge Distillation and On-Policy Distillation}
Knowledge distillation transfers a teacher's soft predictions into a student~\cite{hinton2015distilling}. On a fixed corpus it is off-policy and suffers the same train-inference mismatch as SFT: the student is supervised on prefixes it would not itself produce. On-policy distillation instead trains the student on its own generations, so the teacher gives dense feedback exactly where the student goes~\cite{agarwal2024gkd,lu2025onpolicy}, mirroring imitation learning and its remedy for compounding errors~\cite{ross2011dagger}. These approaches assume a separate, larger teacher; ours requires none, as the same model supplies both roles.

\subsection{Self-Distillation with Privileged Information}
A model can act as its own teacher when given PI unavailable to the student. OPSD conditions the teacher on a ground-truth reference solution and distills its per-token distribution into the student along on-policy rollouts~\cite{zhao2026opsd}, building on the intuition, also exploited by rationalization-based self-training such as STaR~\cite{zelikman2022star}, that explaining a known answer is easier than producing one. A recent analysis dissects how the \emph{structure} of the PI governs whether such self-distillation helps: when PI is instance-specific and absent at test time, the student can only recover a weak PI-marginalized consensus of the PI-conditioned teachers~\cite{zhu2026manyfaces}. A closely related method, On-Policy Context Distillation (OPCD), internalizes an in-context signal, an optimized system prompt or transferable knowledge distilled from a model's own past solution traces, into the student's weights by minimizing the reverse KL to a context-conditioned teacher along the student's own generations~\cite{ye2026onpolicycontextdistillationlanguage}. Its privileged context typically encodes a shared, reusable rule rather than a per-problem label, exactly the regime in which marginalizing over the privilege stays benign; it also remains autoregressive. dOPSD inherits this single-model, privileged-teacher construction but differs in two ways: it operates on diffusion language models, and its privilege is a later state of the student's own denoising trajectory rather than an external reference solution, self-generated and intrinsic to diffusion decoding.

\section{Method}

\begin{figure*}[t]
\centering
\begin{subfigure}[t]{0.49\textwidth}
\centering
\includegraphics[width=\linewidth]{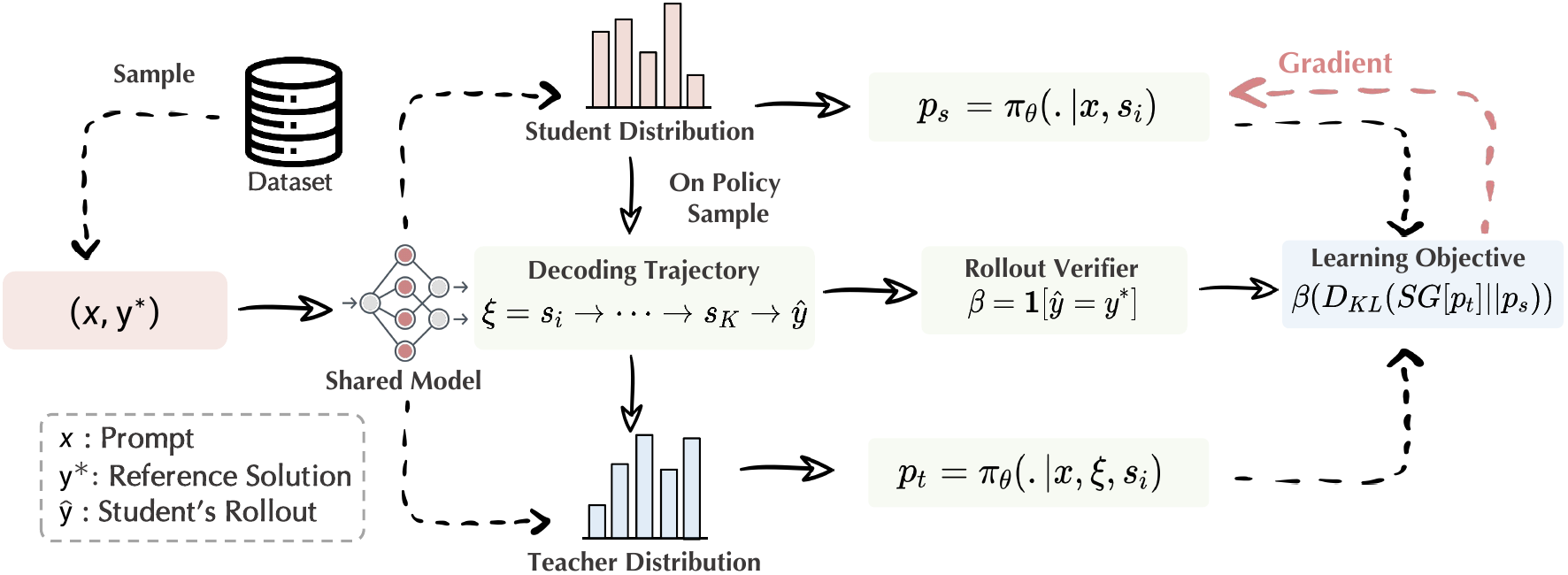}
\caption{Training pipeline.}
\label{fig:overview}
\end{subfigure}
\hfill
\begin{subfigure}[t]{0.48\textwidth}
\centering
\includegraphics[width=\linewidth]{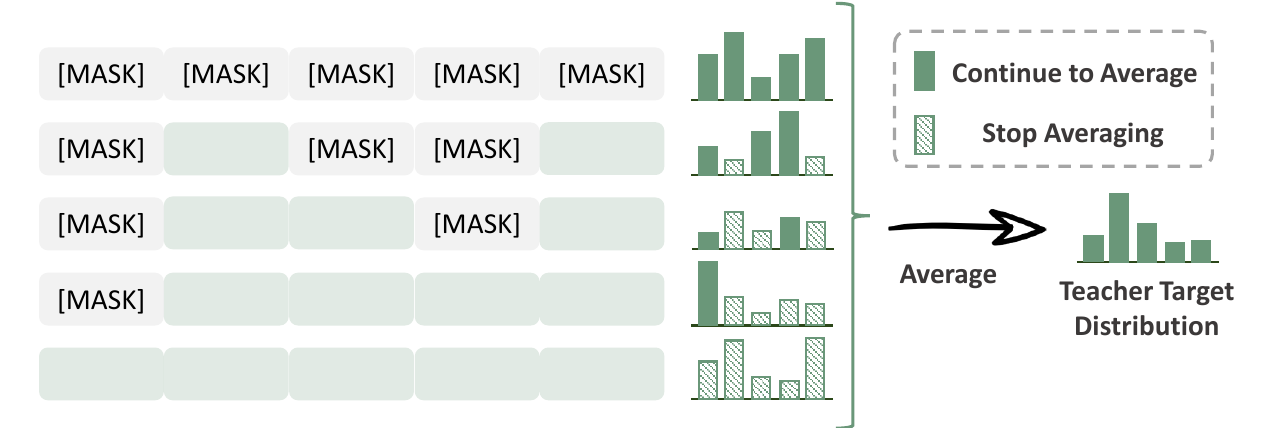}
\caption{Teacher target construction.}
\label{fig:pi}
\end{subfigure}
\caption{Overview of dOPSD. \textbf{(a)} A single shared model plays both the student and the teacher. Given a problem $x$ with reference answer $y^\star$ sampled from the dataset, the model rolls out an on-policy decoding trajectory $\xi=s_i\!\to\!\cdots\!\to\!s_K\!\to\!\hat{y}$. The \emph{student} is scored at an intermediate, still-masked step $s_i$, $p_s=\pi_\theta(\cdot\mid x, s_i)$, while the same model acting as the \emph{teacher} scores the same positions from later, more-decoded states of the trajectory, an on-policy peek-ahead that uses no external solution. A rollout verifier $\beta=\mathbf{1}[\hat{y}=y^\star]$ keeps only correct rollouts, and the student is trained to match the stop-gradient teacher target, with gradients flowing only into the student. \textbf{(b)} Construction of that teacher target at a masked position: the teacher's predictive distribution is averaged over the future decoding steps where the position is \emph{still} masked (solid, the set $\mathcal{T}_i$) and dropped once the position has been decoded (hatched, where the model would merely copy the committed token); the average is the teacher target $\bar{p}_i$ the student is distilled toward.}
\label{fig:dopsd}
\end{figure*}

We first review on-policy self-distillation (OPSD) for autoregressive models (Section~\ref{sec:opsd}). We then introduce how a diffusion language model decodes and analyze why OPSD fails in this setting (Section~\ref{sec:dllm}), and finally present dOPSD (Section~\ref{sec:dopsd}), whose full training pipeline is illustrated in Figure~\ref{fig:overview}.

\subsection{On-Policy Self-Distillation}
\label{sec:opsd}

\paragraph{Setup.} Given a dataset $\mathcal{S}=\{(x_i, y^\star_i)\}_{i=1}^{N}$ with input $x$ and full solution $y^\star$ available only during training (e.g., a reference solution, but possibly a context or system prompt), OPSD post-trains a single model $\pi_\theta$ to perform well when conditioned on $x$ alone. We write $\pi_\theta(\cdot\mid c)$ for the next-token distribution over a vocabulary $\mathcal{V}$ given a context $c$.

\paragraph{Student, teacher, and objective.} From the same parameters $\theta$, OPSD forms a \emph{student} $p_S(\cdot\mid x)\triangleq \pi_\theta(\cdot\mid x)$ that matches the inference-time condition and a \emph{teacher} $p_T(\cdot\mid x,y^\star)\triangleq \pi_\theta(\cdot\mid x,y^\star)$ that also reads the PI; the teacher is stronger only in that it may exploit information the student lacks. To stay on-policy, the student is supervised on its own samples $\hat{y}\sim p_S(\cdot\mid x)$, with both policies scored along this rollout and matched per token,
\begin{equation}
\label{eq:opsd}
\mathcal{L}_{\mathrm{OPSD}}(\theta) =
\mathbb{E}_{(x,y^\star)\sim\mathcal{S}}\,
\mathbb{E}_{\hat{y}\sim p_S(\cdot\mid x)}
\!\left[ \frac{1}{|\hat{y}|} \sum_{n=1}^{|\hat{y}|}
D\!\big( p_T^{\,n} \,\big\|\, p_S^{\,n} \big) \right]\!,
\end{equation}
where $p_T^{\,n}=p_T(\cdot\mid x, y^\star, \hat{y}_{<n})$, $p_S^{\,n}=p_S(\cdot\mid x, \hat{y}_{<n})$, and gradients flow only through the student (the teacher is a stop-gradient target). The divergence $D$ is the forward KL, the reverse KL, or the generalized Jensen--Shannon divergence interpolating between them with a weight $\beta\in[0,1]$,
\begin{equation}
\label{eq:jsd}
\mathrm{JSD}_\beta(p_T \,\|\, p_S) = \beta\,\mathrm{KL}(p_T \,\|\, m) + (1-\beta)\,\mathrm{KL}(p_S \,\|\, m),
\end{equation}
with mixture $m=\beta p_T+(1-\beta)p_S$. This yields dense, on-policy, token-level supervision with no external teacher or reward model.

\subsection{Why OPSD Fails for Diffusion Language Models}
\label{sec:dllm}

\paragraph{Decoding as a denoising trajectory.} A dLLM generates a length-$L$ completion for a prompt $x$ by reverse denoising. Starting from a fully masked sequence, it repeatedly predicts the clean token at each masked position and commits only the most confident ones, freezing them for the rest of decoding. Over $K$ steps this traces a sequence of partial states
\begin{equation}
\label{eq:traj}
s_0 \to s_1 \to \cdots \to s_K = \hat{y},
\end{equation}
where $s_0$ is fully masked, $s_K=\hat{y}$ is the finished completion, and each intermediate $s_k$ equals $\hat{y}$ with its not-yet-decoded positions held at $\langle\mathrm{mask}\rangle$. Let $M_k\subseteq\{1,\dots,L\}$ be the positions still masked at step $k$. For the decoders we consider, a token is kept once decoded, so the masked set shrinks monotonically, $M_0\supseteq M_1\supseteq\cdots\supseteq M_K=\emptyset$; our method requires only that each $s_k$ is a genuine intermediate state visited during decoding. At any state the model provides a masked-prediction distribution $\pi_\theta(\cdot\mid x, s_k)_i$ over $\mathcal{V}$ at each masked position $i\in M_k$. Applying OPSD here meets two obstacles: one inherited from OPSD itself, the other specific to how a dLLM decodes. We examine each in turn: the PI collapses into a PI-free policy, and the random masking OPSD imposes drifts off the model's actual decoding order.

\begin{figure*}[t]
\centering
\begin{minipage}{0.93\textwidth}
\centering

\begin{subfigure}[t]{0.32\linewidth}
\centering
\includegraphics[width=\linewidth]{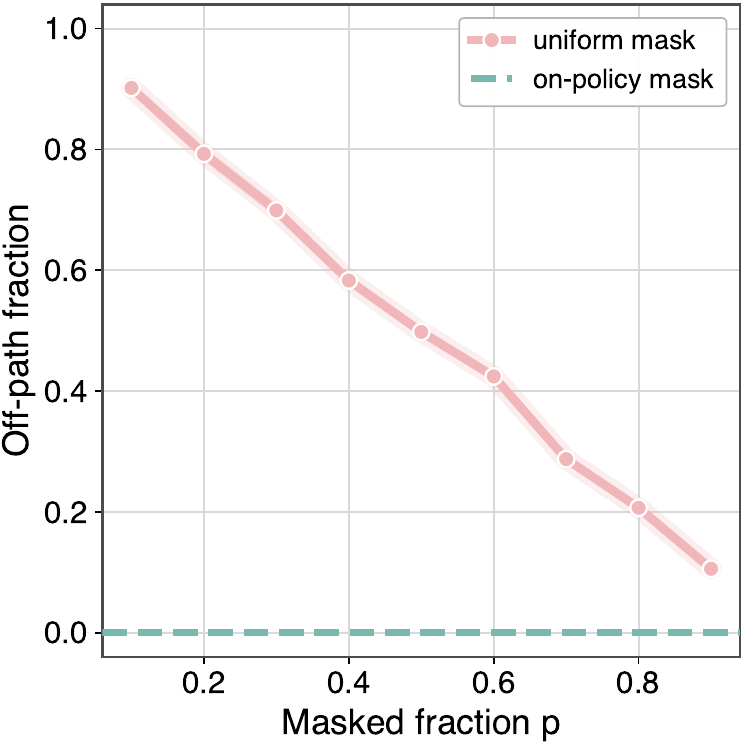}
\caption{Off-path fraction of a uniform mask.}
\label{fig:offpath}
\end{subfigure}
\hfill
\begin{subfigure}[t]{0.32\linewidth}
\centering
\includegraphics[width=\linewidth]{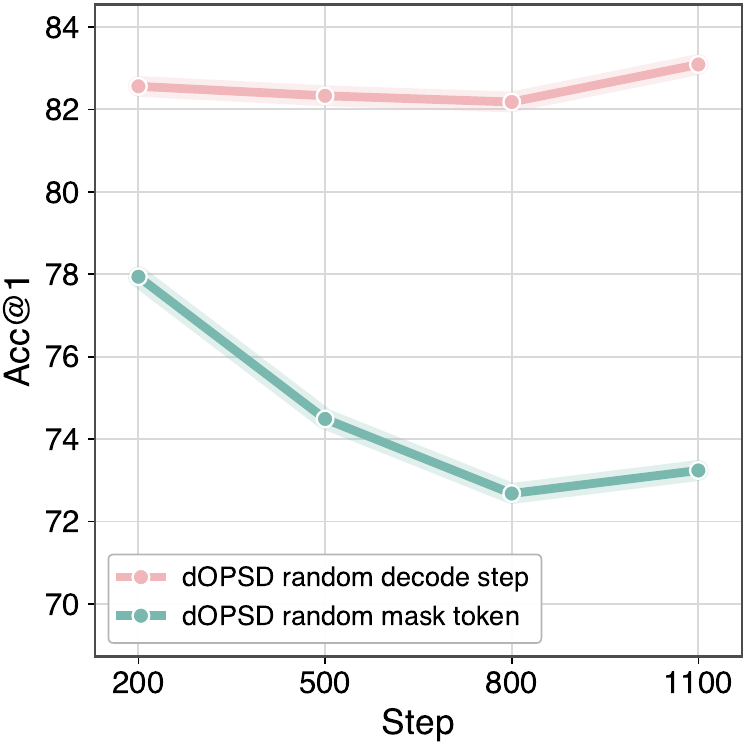}
\caption{Random mask tokens vs.\ decode steps.}
\label{fig:decodemask}
\end{subfigure}
\hfill
\begin{subfigure}[t]{0.32\linewidth}
\centering
\includegraphics[width=\linewidth]{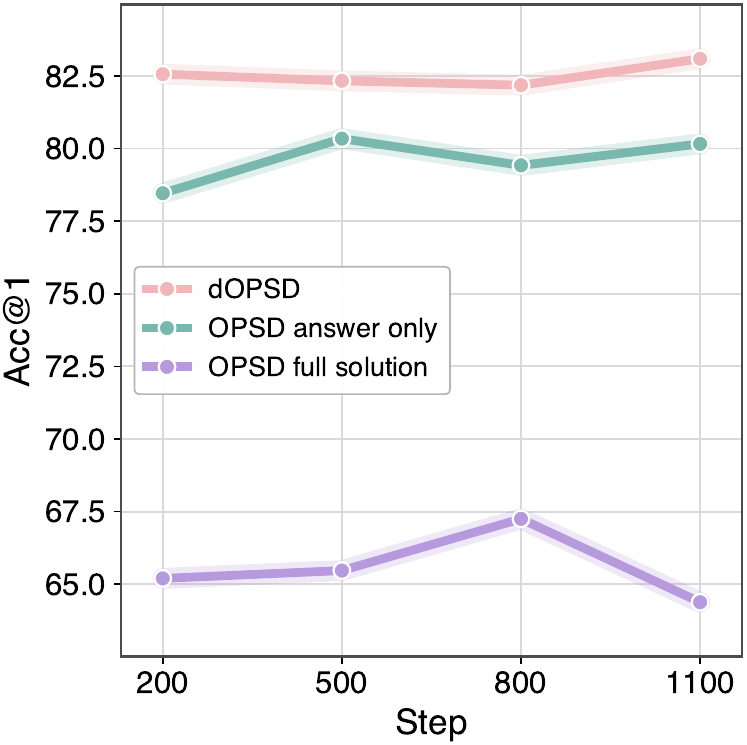}
\caption{Performance of OPSD variants vs.\ dOPSD.}
\label{fig:gsm8k}
\end{subfigure}
\end{minipage}
\caption{OPSD cannot be applied directly to diffusion language models (Dream-7B-Instruct). \textbf{(a)} The fraction of a uniformly masked subset that lands above the on-policy decoding frontier, easy tokens the model would already have committed, is large at small mask fractions $p$, whereas a mask taken from a genuine decoding step (blue) is zero by construction. \textbf{(b)} Scoring the student at random mask tokens of the finished rollout (off-path) degrades GSM8K accuracy over training, while scoring it at genuine random decode steps stays high. \textbf{(c)} Ported directly to a dLLM, both OPSD variants train to substantially lower accuracy than our trajectory-based dOPSD, and conditioning the teacher on the full reference solution collapses performance.}
\label{fig:obstacles}

\end{figure*}

\paragraph{The privilege is external (PI-free collapse).} The benefits of Section~\ref{sec:opsd} all flow from the privileged teacher, which is also where the method breaks. The student is never conditioned on $y^\star$, so minimizing the expected divergence over the PI distribution $\mathcal{D}(\cdot\mid x)$ drives it toward a single \emph{PI-free} policy~\cite{zhu2026manyfaces}, the normalized geometric mean of the per-PI teachers,
\begin{equation}
\label{eq:pifree}
p_S^\star(y \mid x) =
\frac{\exp\!\big(\mathbb{E}_{y^\star\sim\mathcal{D}(\cdot\mid x)}\,\log p_T(y\mid x, y^\star)\big)}
{\sum_{y'}\exp\!\big(\mathbb{E}_{y^\star\sim\mathcal{D}(\cdot\mid x)}\,\log p_T(y'\mid x, y^\star)\big)}.
\end{equation}
A response survives this average only when it is supported across values of $y^\star$. When $y^\star$ encodes a shared latent rule (a system prompt or alignment preference), the average compresses it into a reusable behavior; when $y^\star$ is \emph{instance-specific}, as a per-problem reference solution is, each input induces its own incompatible teacher and the student collapses onto a weak consensus far worse than any single teacher. Mathematical reasoning, our target, is squarely this instance-specific case, so OPSD's gains evaporate exactly where we want them. Figure~\ref{fig:gsm8k} previews this collapse on Dream: ported directly to a dLLM, OPSD trains to substantially lower GSM8K accuracy than our trajectory-based dOPSD (Section~\ref{sec:dopsd}), and, exactly as the PI-free account predicts, conditioning the teacher on the richer, fully instance-specific reference solution is dramatically worse than an answer-only teacher, trailing by roughly $15$ points.

\paragraph{The noise is off the decoding path.} The second obstacle is specific to how a dLLM decodes. The autoregressive recipe has no notion of masking: it scores every token of the rollout in one pass. The natural port is to roll out a completion $\hat{y}$, re-mask a uniformly random subset of its positions, and match teacher and student at those positions. That random corruption, however, is not how a dLLM decodes. At inference the masked positions of a partial state follow a confidence-ordered, easy-to-hard schedule, and their visible context is whatever the model has already committed; a uniformly random mask over the finished rollout produces partial states the model never visits. Figure~\ref{fig:offpath} measures this drift directly: we roll out completions, recover the confidence-ordered decoding frontier of each, and ask what fraction of a uniformly masked subset lands on tokens the model would already have committed, those above the frontier. At a $10\%$ mask roughly $90\%$ of the masked positions are such already-decoded-easy tokens, and even at a $50\%$ mask about half are; masking instead the not-yet-decoded positions of a genuine decoding step puts zero mass above the frontier by construction. The supervision then falls off the model's own decoding path, so even setting the PI problem aside the signal is mismatched with inference. Training bears out the cost: Figure~\ref{fig:decodemask} shows that scoring the student at random masked tokens of the finished rollout steadily degrades GSM8K accuracy over training, whereas scoring it at genuine random decode steps stays high and stable. Both failures share one cause: the privilege and the noise are imposed from outside the model rather than drawn from its own generation.

% \begin{figure*}[t]
% \centering
% \begin{subfigure}[t]{0.32\textwidth}
% \centering
% \includegraphics[width=\linewidth]{Figures/offpath_fraction}
% \caption{Off-path fraction of a uniform mask.}
% \label{fig:offpath}
% \end{subfigure}
% \hfill
% \begin{subfigure}[t]{0.32\textwidth}
% \centering
% \includegraphics[width=\linewidth]{Figures/decode_vs_mask}
% \caption{Random mask tokens vs.\ decode steps.}
% \label{fig:decodemask}
% \end{subfigure}
% \hfill
% \begin{subfigure}[t]{0.32\textwidth}
% \centering
% \includegraphics[width=\linewidth]{Figures/gsm8k_acc}
% \caption{OPSD variants vs.\ dOPSD.}
% \label{fig:gsm8k}
% \end{subfigure}
% \caption{OPSD cannot be applied directly to diffusion language models (Dream-7B-Instruct). \textbf{(a)} The fraction of a uniformly masked subset that lands above the on-policy decoding frontier, easy tokens the model would already have committed, is large at small mask fractions $p$, whereas a mask taken from a genuine decoding step (blue) is zero by construction. \textbf{(b)} Scoring the student at random mask tokens of the finished rollout (off-path) degrades GSM8K accuracy over training, while scoring it at genuine random decode steps stays high. \textbf{(c)} Ported directly to a dLLM, both OPSD variants train to substantially lower accuracy than our trajectory-based dOPSD, and conditioning the teacher on the full reference solution collapses performance.}
% \label{fig:obstacles}
% \end{figure*}

\subsection{dOPSD}
\label{sec:dopsd}

dOPSD removes both obstacles by reading the student's noisy state, the teacher's privilege, and the noise itself off a single object: the student's own decoding trajectory (Eq.~\ref{eq:traj}). For each prompt the student decodes a completion and we record its trajectory $s_0,\dots,s_K$; everything below is taken from this trajectory, with nothing masked or supplied from outside it.

\paragraph{Student state: a real unmasking step.} Instead of re-masking the finished rollout at random, we sample an intermediate decoding step $k$ uniformly at random among the steps whose masked fraction $|M_k|/L$ is still above a threshold $\tau$, so the state remains substantially masked and offers enough positions to score. The student is then evaluated exactly as during decoding: at the genuinely masked positions $M_k$, conditioned on the real partial state $s_k$. The noise is thus a true intermediate step of the model's own denoising, keeping the supervised states on the inference path.

\paragraph{Teacher privilege: the remaining trajectory.} The teacher scores the same positions $M_k$, but with the extra context that the student's \emph{own later decoding} supplies. For a masked position $i\in M_k$, let
\begin{equation}
\label{eq:tset}
\mathcal{T}_i = \{\, t : k < t \le K,\ i\in M_t \,\}
\end{equation}
collect the future steps at which $i$ is \emph{still} masked. At each such step the model has committed more of the surrounding tokens, so a masked-prediction pass on the later state $s_t$ yields a better-informed distribution over position $i$ than the student had at step $k$. We take the teacher target $\bar{p}_i$ to be the average of these predictions (Figure~\ref{fig:pi}),
\begin{equation}
\label{eq:teacher}
\bar{p}_i = \frac{1}{|\mathcal{T}_i|}\sum_{t\in\mathcal{T}_i} \pi_\theta(\cdot\mid x, s_t)_i .
\end{equation}
Averaging over the entire remaining masked lifetime of $i$, rather than a single later snapshot, gathers every step at which the teacher is genuinely better informed \emph{while $i$ is still predicted rather than copied}: steps at which $i$ has already been decoded are excluded from $\mathcal{T}_i$, since there the model would merely read back the committed token instead of predicting it. A position committed at the very next step ($\mathcal{T}_i=\emptyset$) has no such genuine future prediction, and we fall back to the endpoint teacher $\pi_\theta(\cdot\mid x, \hat{y})_i$. All of this privilege is produced by the model's own decoding, so it is on-policy, self-generated, and uses no reference solution, which is exactly what the PI-free analysis calls for.

\paragraph{Objective.} We distill the averaged teacher $\bar{p}_i$ into the present-step student at the masked positions with the token-level generalized Jensen--Shannon divergence of Eq.~\eqref{eq:jsd},
\begin{equation}
\label{eq:dopsd}
\ell(x,\hat{y}) = \frac{1}{|M_k|}\sum_{i\in M_k}
\mathrm{JSD}_\beta\!\Big( \mathrm{sg}\big[\,\bar{p}_i\,\big] \,\big\|\, \pi_\theta(\cdot\mid x, s_k)_i \Big),
\end{equation}
where $\mathrm{sg}[\cdot]$ stops the gradient on the teacher target. The sum runs only over $M_k$: positions already decoded by step $k$ are fixed context and incur no loss, matching the masked-prediction objective the model is trained with. Both terms are ordinary masked-prediction passes, so dOPSD adds no architectural change; its only extra cost is one teacher forward per distinct remaining step in Eq.~\eqref{eq:teacher}.

\paragraph{Verifying the rollout.} A self-generated trajectory carries no guarantee of correctness, and distilling a confidently wrong rollout would only reinforce the error. We therefore verify each finished completion $\hat{y}$ against the ground-truth final answer and keep only correct rollouts in the loss; trajectories with a wrong answer are discarded and contribute no gradient. This verification is not essential to the method: when the dataset provides no reference answer, dOPSD can simply distill from every rollout, which we confirm in Section~\ref{sec:ablation}. The objective averages $\ell$ over the verified-correct rollouts,
\begin{equation}
\label{eq:ldopsd}
\mathcal{L}_{\mathrm{dOPSD}}(\theta) =
\mathbb{E}_{x}\,\mathbb{E}_{\hat{y}\sim \pi_\theta(\cdot\mid x)}
\big[\, \mathbf{1}[\hat{y}\ \text{correct}]\;\ell(x,\hat{y}) \,\big].
\end{equation}
The verifier reads only the short final answer, not a reference solution, so the teacher's privilege stays entirely trajectory-derived. Algorithm~\ref{alg:dopsd} summarizes the full procedure.

\begin{algorithm}[t]
\caption{dOPSD Training}
\label{alg:dopsd}
\textbf{Require}: dataset $\mathcal{S}=\{(x_i, a_i)\}_{i=1}^{N}$ with final answers $a_i$; diffusion LM $\pi_\theta$; mask threshold $\tau$
\begin{algorithmic}[1]
\WHILE{not converged}
\STATE Sample a minibatch $\mathcal{B}\subset\mathcal{S}$
\FOR{all $(x,a)\in\mathcal{B}$}
\STATE Roll out $\hat{y}$ with its trajectory $(s_0,\dots,s_K)\sim \pi_\theta(\cdot\mid x)$
\IF{$\hat{y}$ disagrees with the gold answer $a$}
\STATE $\ell(x)\leftarrow 0$ \COMMENT{discard wrong rollout}
\ELSE
\STATE Sample step $k$ uniformly from $\{k': |M_{k'}|/L > \tau\}$
\STATE Build teacher targets $\{\bar{p}_i\}_{i\in M_k}$ over the remaining masked steps (Eq.~\eqref{eq:teacher})
\STATE Score $M_k$ via Eq.~\eqref{eq:dopsd} to obtain $\ell(x)$
\ENDIF
\ENDFOR
\STATE $\mathcal{L}_{\mathrm{dOPSD}}\!\leftarrow\! \frac{1}{|\mathcal{B}|}\sum_{(x,a)\in\mathcal{B}}\ell(x)$; update $\theta$
\ENDWHILE
\end{algorithmic}
\end{algorithm}

\section{Experiments}
\label{sec:experiments}

\subsection{Experimental Setup}

\paragraph{Models.} We build on two open, instruction-tuned diffusion language models, Dream-7B-Instruct~\cite{ye2025dream} and LLaDA-8B-Instruct~\cite{nie2025llada}, and apply every method as a post-training stage on top of them.

\paragraph{Training data.} We focus on mathematical reasoning and train on MixChain-Z-PRM12K, a corpus of competition-style problems paired with chain-of-thought solutions and verifiable final answers. To probe generalization, we additionally evaluate the trained models, without any further fine-tuning, on out-of-distribution code generation.

\paragraph{Baselines.} We compare dOPSD against four post-training methods trained on the same data and base models, together with the untuned base model: (i) \textbf{SFT}, supervised fine-tuning on the reference solutions, an off-policy imitation baseline; (ii) \textbf{GRPO}~\cite{zhao2025d1}, group-relative policy optimization with a verifiable final-answer reward, adapted to diffusion decoding; (iii) \textbf{OPSD (answer-only)}, the on-policy self-distillation port whose teacher is conditioned only on the final answer as PI; and (iv) \textbf{OPSD (full-solution)}, the same port whose teacher is conditioned on the complete reference solution. The two OPSD variants are exactly the naive transfers analyzed in Section~\ref{sec:dllm}.

\paragraph{Training details.} All methods use parameter-efficient LoRA \cite{hu2022lora} fine-tuning with rank $32$ and scaling $\alpha=32$, applied to the \texttt{q\_proj}, \texttt{k\_proj}, \texttt{v\_proj}, \texttt{o\_proj}, \texttt{gate\_proj}, \texttt{up\_proj}, and \texttt{down\_proj} projections. Training runs on $8$ GPUs with a per-device mini-batch size of $4$, optimized with AdamW at a peak learning rate of $2\times10^{-5}$ for $3$ epochs; for every method we report the best checkpoint. For dOPSD we set the trajectory mask threshold to $\tau=0.5$, so the student is scored at a decoding step whose answer region is still at least half masked.

\paragraph{Evaluation.} We measure in-domain mathematical reasoning on GSM8K \cite{gsm8k} and MATH500 \cite{math500}, and out-of-distribution code generation on HumanEval \cite{humaneval} and MBPP \cite{mbpp}, reporting accuracy on the math benchmarks and pass@1 on the code benchmarks.

\subsection{Main Results}

\begin{table*}[t]
\centering
\begin{tabular}{l cc c cc c}
\toprule
& \multicolumn{3}{c}{In-Domain: Math Reasoning} & \multicolumn{3}{c}{OOD: Code Generation} \\
\cmidrule(lr){2-4} \cmidrule(lr){5-7}
Method & GSM8K & MATH500 & Avg. & HumanEval & MBPP & Avg. \\
\midrule
\multicolumn{7}{l}{\textit{Base Model: Dream-7B-Instruct}} \\
Base & \underline{81.41} & \underline{38.97} & \underline{60.19} & \underline{52.54} & \underline{57.43} & \underline{54.99} \\
SFT & 80.44 & 36.49 & 58.47 & 48.27 & 56.94 & 52.61 \\
GRPO & 81.28 & 37.86 & 59.57 & 50.24 & 57.32 & 53.78 \\
OPSD (answer-only) & 80.34 & 37.23 & 58.79 & 47.28 & 55.49 & 51.39 \\
OPSD (full-solution) & 67.25 & 35.72 & 51.49 & 40.86 & 49.28 & 45.07 \\
dOPSD (Ours) & \textbf{83.04} & \textbf{42.20} & \textbf{62.62} & \textbf{56.71} & \textbf{58.49} & \textbf{57.60} \\
\midrule
\multicolumn{7}{l}{\textit{Base Model: LLaDA-8B-Instruct}} \\
Base & 71.23 & 31.24 & 51.24 & \underline{36.12} & \underline{38.45} & \underline{37.29} \\
SFT & 69.45 & 30.12 & 49.79 & 34.27 & 35.24 & 34.76 \\
GRPO & 70.45 & \underline{32.77} & \underline{51.61} & 35.78 & 36.44 & 36.11 \\
OPSD (answer-only) & \underline{71.47} & 31.45 & 51.46 & 35.54 & 36.24 & 35.89 \\
OPSD (full-solution) & 57.86 & 24.34 & 41.10 & 30.67 & 28.78 & 29.73 \\
dOPSD (Ours) & \textbf{72.87} & \textbf{36.00} & \textbf{54.44} & \textbf{39.63} & \textbf{39.54} & \textbf{39.59} \\
\bottomrule
\end{tabular}
\caption{Performance of dOPSD against supervised, reinforcement-learning, and on-policy self-distillation baselines on Dream-7B-Instruct and LLaDA-8B-Instruct. All methods are trained on the mathematical-reasoning corpus; code generation is evaluated out of distribution. The best result in each column is in \textbf{bold} and the second best is \underline{underlined}.}
\label{tab:main}
\end{table*}

\paragraph{dOPSD is the only method that improves the base model.} Across both Dream-7B-Instruct and LLaDA-8B-Instruct, dOPSD attains the best score on every benchmark, in-domain and out-of-distribution (Table~\ref{tab:main}). On Dream it raises GSM8K from $81.41$ to $83.04$ and MATH500 from $38.97$ to $42.20$; on LLaDA it improves MATH500 by $4.76$ points ($31.24\!\to\!36.00$) and GSM8K by $1.64$. Every baseline, by contrast, matches or falls below the base model on nearly every benchmark and improves it by at most a marginal amount, whereas dOPSD improves it across the board.

\paragraph{Supervised, RL, and naive OPSD baselines degrade reasoning.} SFT falls below the base model on all four tasks for both backbones, consistent with the exposure bias of off-policy imitation. GRPO stays at or below the base model as well (its only gain is $+1.53$ on LLaDA MATH500), reflecting the sparse, sequence-level reward and the difficulty of policy optimization without a tractable sequence likelihood. The distillation ports fare no better: OPSD (answer-only) hovers around the base model, while OPSD (full-solution) collapses, losing $14.2$ GSM8K points on Dream ($81.41\!\to\!67.25$) and $13.4$ on LLaDA ($71.23\!\to\!57.86$).

\paragraph{The OPSD ports confirm the PI-free hypothesis.} These patterns are precisely what our analysis in Section~\ref{sec:dllm} predicts, and we read them as direct support for the PI-free hypothesis. First, because the student can never condition on the instance-specific reference solution at inference, OPSD can only distill the weak PI-marginalized consensus of Eq.~\eqref{eq:pifree}; accordingly, OPSD (answer-only) never rises meaningfully above the base model. Second, the failure should \emph{worsen} as the PI becomes richer and more instance-specific, yielding more sharply incompatible per-problem teachers. The data bear this out: the ordering OPSD (full-solution) $<$ OPSD (answer-only) holds on every benchmark and both backbones, with the full-solution teacher collapsing reasoning outright. dOPSD avoids the failure mode entirely by sourcing the teacher's privilege from the model's own trajectory rather than an external label.

\paragraph{Gains transfer out of distribution.} Although trained only on mathematics, dOPSD is the sole method that also improves code generation, lifting HumanEval by $4.17$ points on Dream ($52.54\!\to\!56.71$) and $3.51$ on LLaDA ($36.12\!\to\!39.63$), whereas SFT and both OPSD ports lose ground. The trajectory-derived signal thus strengthens reasoning without the catastrophic forgetting of the supervised and PI-conditioned baselines.

\subsection{Ablation Studies}
\label{sec:ablation}

\begin{figure}[t]
    \centering
    \begin{subfigure}[t]{0.46\columnwidth}
        \centering
        \includegraphics[width=\linewidth]{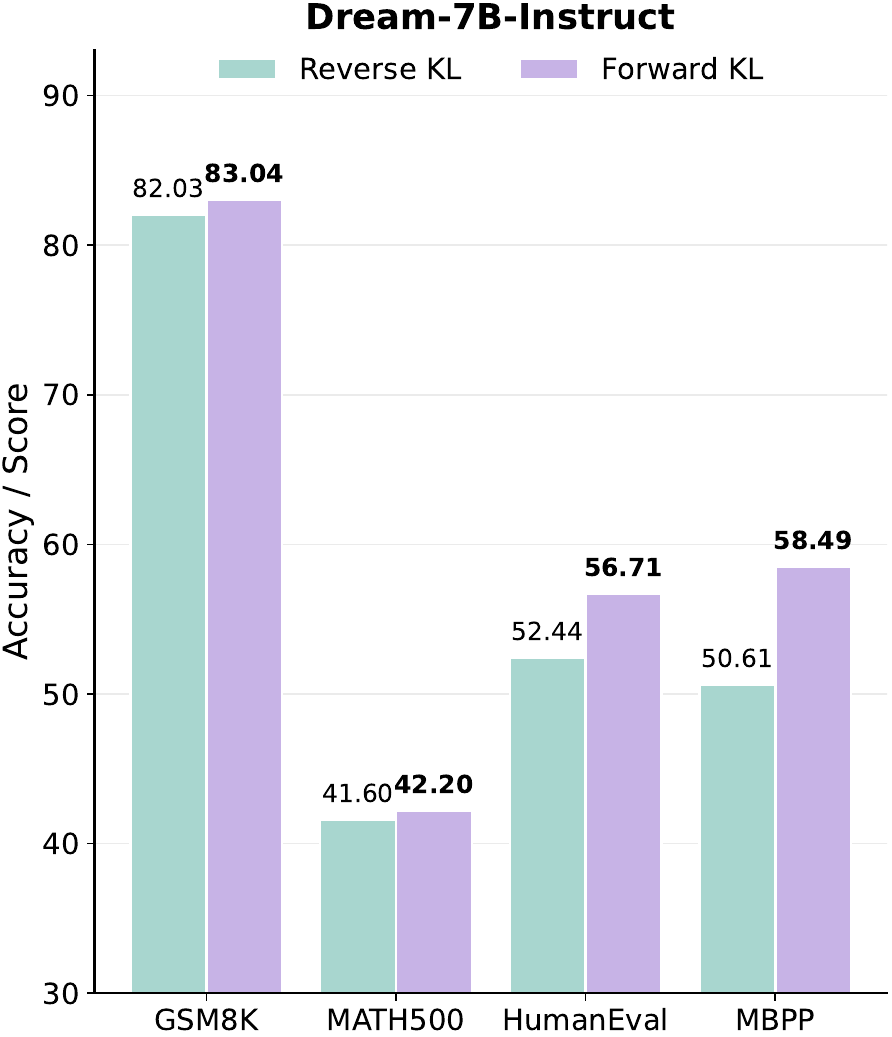}
        \caption{Dream-7B-Instruct}
        \label{fig:kl_dream}
    \end{subfigure}
    \hfill
    \begin{subfigure}[t]{0.46\columnwidth}
        \centering
        \includegraphics[width=\linewidth]{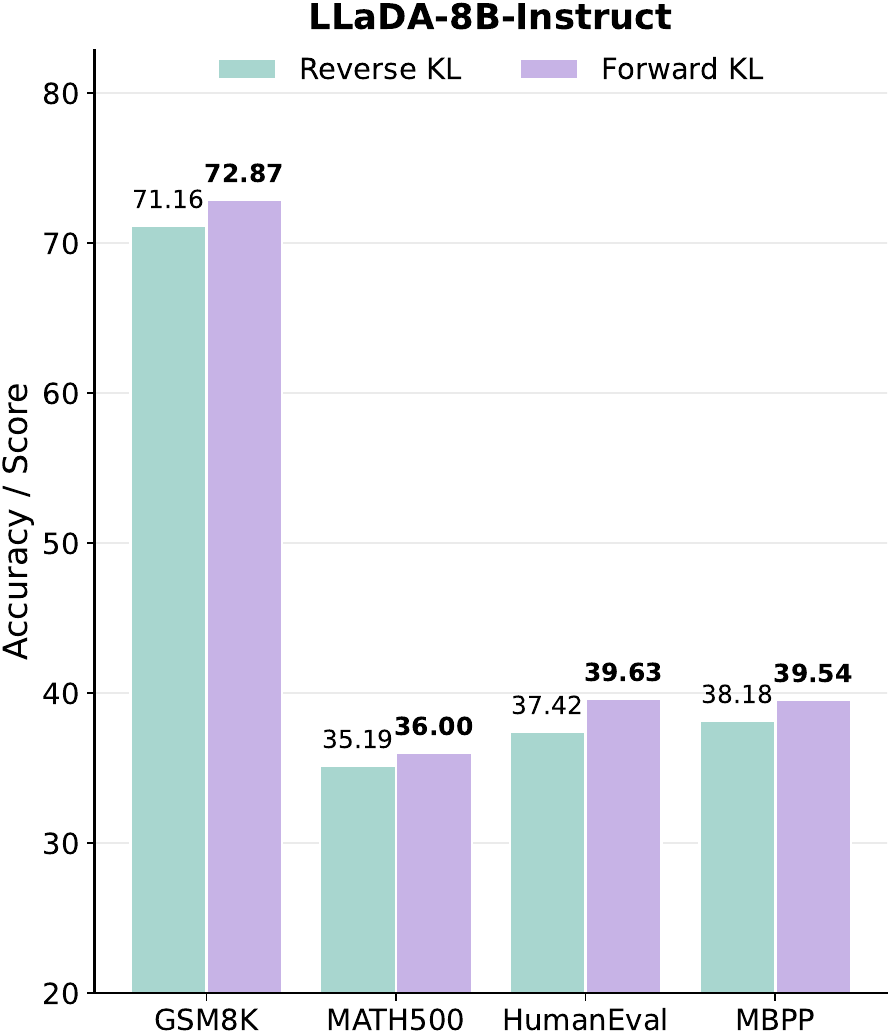}
        \caption{LLaDA-8B-Instruct}
        \label{fig:kl_llada}
    \end{subfigure}
    \caption{Effect of the distillation divergence in dOPSD: forward KL ($\beta\!\to\!0$, default) versus reverse KL ($\beta\!\to\!1$), with all other settings fixed. Best results in each benchmark are highlighted in \textbf{bold}.}
    \label{fig:kl}
\end{figure}

\paragraph{Forward versus reverse KL.} dOPSD distills with the generalized Jensen--Shannon divergence of Eq.~\eqref{eq:dopsd}, whose two limiting cases are forward KL ($\beta\!\to\!0$), our default, and reverse KL ($\beta\!\to\!1$). Figure~\ref{fig:kl} compares them under otherwise identical settings. Forward KL is the stronger choice, winning on every task-backbone cell for both models. The advantage is consistent across both domains; for instance, LLaDA HumanEval rises from $37.42$ to $39.63$ and Dream MBPP from $50.61$ to $58.49$. We attribute this to the direction of the divergence: forward KL is mass-covering, so the student is trained to reproduce the entire peek-ahead teacher distribution and absorb all of its trajectory-derived privileged signal, whereas the mode-seeking reverse KL lets the student concentrate on a single mode and discard part of that signal. We therefore use forward KL throughout.

% \begin{table}[t]
% \centering
% \setlength{\tabcolsep}{4pt}
% \begin{tabular}{l cccc}
% \toprule
% & GSM8K & MATH500 & HumanEval & MBPP \\
% \midrule
% \multicolumn{5}{l}{\textit{Dream-7B-Instruct}} \\
% Reverse KL & 82.03 & 41.60 & 52.44 & 50.61 \\
% Forward KL & \textbf{83.04} & \textbf{42.20} & \textbf{56.71} & \textbf{58.49} \\
% \midrule
% \multicolumn{5}{l}{\textit{LLaDA-8B-Instruct}} \\
% Reverse KL & 71.16 & 35.19 & 37.42 & 38.18 \\
% Forward KL & \textbf{72.87} & \textbf{36.00} & \textbf{39.63} & \textbf{39.54} \\
% \bottomrule
% \end{tabular}
% \caption{Effect of the distillation divergence in dOPSD: forward KL ($\beta\!\to\!0$, default) versus reverse KL ($\beta\!\to\!1$), with all other settings fixed. Best in each column in \textbf{bold}.}
% \label{tab:kl}
% \end{table}

\paragraph{Teacher horizon: how many future steps.} The teacher target $\bar{p}_i$ (Eq.~\eqref{eq:teacher}) averages the teacher's prediction over the future steps at which position $i$ is still masked. We ablate how far into the trajectory this window reaches by capping it at $0$, $50$, $100$, or $200$ future steps, versus the full remaining trajectory (our default), on Dream (Table~\ref{tab:horizon}). Performance improves as the teacher looks further ahead: with no future step the teacher carries no peek-ahead privilege and trails the base model badly (GSM8K $70.44$), and each wider horizon adds more genuinely-decoded context that sharpens the target. Averaging over the full remaining trajectory is best on every benchmark by a wide margin (e.g., GSM8K $83.04$ vs.\ $78.14$ at $200$ steps, and MBPP $58.49$ vs.\ $48.92$), confirming that the privileged signal is strongest when the teacher aggregates the model's entire subsequent decoding rather than a truncated look-ahead. We therefore use the full horizon throughout.

\begin{table}[t]
\centering
\setlength{\tabcolsep}{4pt}
\begin{tabular}{l cccc}
\toprule
Future steps & GSM8K & MATH500 & HumanEval & MBPP \\
\midrule
$0$ & 70.44 & 30.24 & 35.46 & 40.24 \\
$50$ & 74.26 & 34.89 & 32.61 & 39.45 \\
$100$ & 78.42 & 33.78 & 40.22 & 42.75 \\
$200$ & 78.14 & 35.48 & 46.45 & 48.92 \\
Full & \textbf{83.04} & \textbf{42.20} & \textbf{56.71} & \textbf{58.49} \\
\bottomrule
\end{tabular}
\caption{Effect of the teacher horizon on Dream-7B-Instruct: the number of future trajectory steps averaged into the teacher target $\bar{p}_i$ (Eq.~\eqref{eq:teacher}). ``Full'' averages over the entire remaining masked trajectory (our default). Best in each column in \textbf{bold}.}
\label{tab:horizon}
\end{table}

\paragraph{Trajectory mask threshold.} The threshold $\tau$ controls how masked the chosen student step must be: only decoding steps whose masked fraction exceeds $\tau$ are eligible, so a small $\tau$ scores the student on nearly finished states and a large $\tau$ on almost-empty ones. Table~\ref{tab:tau} varies $\tau$ on Dream. A moderate $\tau=0.5$ is best, with the highest average over the four tasks ($60.11$) and the top score on every benchmark except HumanEval. It surpasses both $\tau=0.25$ ($53.63$ average), which leaves too few masked positions to learn from, and $\tau=0.75$ ($59.49$), whose targets are too heavily masked to predict reliably. The gain holds both in domain (GSM8K, MATH500) and out of distribution (MBPP), with $\tau=0.75$ only marginally ahead on HumanEval. We use $\tau=0.5$ throughout.

\begin{table}[t]
\centering
\setlength{\tabcolsep}{4pt}
\begin{tabular}{l ccccc}
\toprule
$\tau$ & GSM8K & MATH500 & HumanEval & MBPP & Avg. \\
\midrule
$0.25$ & 79.25 & 36.57 & 50.42 & 48.28 & 53.63 \\
$0.50$ & \textbf{83.04} & \textbf{42.20} & 56.71 & \textbf{58.49} & \textbf{60.11} \\
$0.75$ & 81.67 & 41.60 & \textbf{57.41} & 57.28 & 59.49 \\
\bottomrule
\end{tabular}
\caption{Effect of the trajectory mask threshold $\tau$ on Dream-7B-Instruct. ``Avg.'' is the mean over the four benchmarks; $\tau=0.5$ (default) is best. Best in each column in \textbf{bold}.}
\label{tab:tau}
\end{table}

\begin{figure}[t]
    \centering
    \includegraphics[width=\columnwidth]{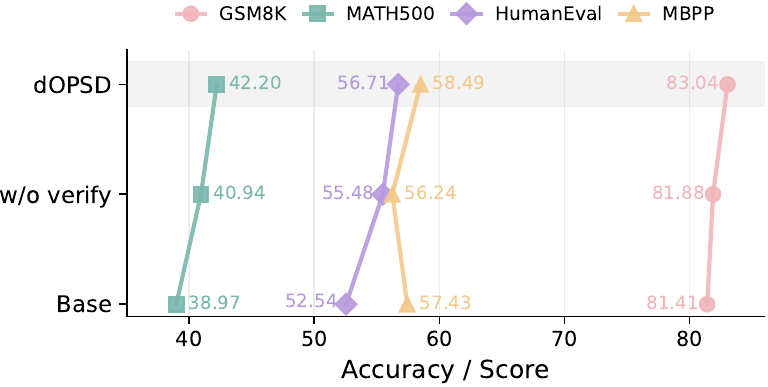}
    \caption{Effect of removing rollout verification on Dream-7B-Instruct. Distilling from \emph{all} rollouts (\emph{w/o verify}, no reference answer) still improves over the base model, while the verifier adds a further gain.}
    \label{fig:verify}
\end{figure}

\paragraph{Training without rollout verification.} dOPSD uses the gold answer only to discard incorrect rollouts in Eq.~\eqref{eq:ldopsd}; when no reference answer is available, it can instead distill from every rollout. Figure~\ref{fig:verify} ablates this on Dream. Even without verification, dOPSD still improves over the base model, raising its four-task average from $57.59$ to $58.64$ and gaining on GSM8K, MATH500, and HumanEval (for example MATH500 $38.97\!\to\!40.94$), with only a small drop on MBPP. Verification adds a further boost, to a $60.11$ average, but it is not what makes the method work: the gains stem from the trajectory-derived privileged signal, which needs no answer supervision. dOPSD therefore applies even to datasets without reference answers, unlike the SFT and OPSD baselines, which require reference solutions.

% \begin{table}[t]
% \centering
% \setlength{\tabcolsep}{4pt}
% \begin{tabular}{l cccc}
% \toprule
% & GSM8K & MATH500 & HumanEval & MBPP \\
% \midrule
% Base & 81.41 & 38.97 & 52.54 & \underline{57.43} \\
% w/o verify & \underline{81.88} & \underline{40.94} & \underline{55.48} & 56.24 \\
% dOPSD & \textbf{83.04} & \textbf{42.20} & \textbf{56.71} & \textbf{58.49} \\
% \bottomrule
% \end{tabular}
% \caption{Effect of removing rollout verification on Dream-7B-Instruct. Distilling from \emph{all} rollouts (\emph{w/o verify}, no reference answer) still improves over the base model; the verifier adds a further gain. The best result in each column is highlighted in \textbf{bold}, while the second-best result is \underline{underlined}.}
% \label{tab:verify}
% \end{table}

\section{Conclusion}

We presented dOPSD, an on-policy self-distillation method that adapts the privileged-teacher principle of OPSD to diffusion language models. We first identified why the autoregressive recipe does not transfer: its PI is an external, instance-specific reference solution absent at inference, so OPSD collapses onto a weak PI-free consensus, and its random masking lands off the model's confidence-ordered decoding path. dOPSD resolves both issues by drawing the teacher's privilege and the student's noise from one source, the student's own denoising trajectory: the student is scored at a genuine intermediate step while the same model, as teacher, scores the same positions from later, more-decoded steps, an honest peek-ahead needing no reference solution. On Dream-7B-Instruct and LLaDA-8B-Instruct, dOPSD is the only method to improve the base model across the board, strengthening in-domain mathematical reasoning and, despite training only on mathematics, out-of-distribution code generation, while supervised, RL, and naive OPSD baselines stagnate or collapse. Ablations confirm the gains stem from the trajectory-derived signal itself and hold even without rollout verification, so dOPSD applies to datasets with no reference answers. We hope that casting the denoising trajectory as a source of PI offers a useful, label-efficient perspective for post-training diffusion language models.

\newpage
\bibliography{aaai2027}

\newpage
\appendix

\section{Dataset Details}
\label{app:datasets}

This appendix expands on the training corpus and evaluation benchmarks summarized in the Experimental Setup, including their sources, sizes, preprocessing, and scoring protocols.

\subsection{Training Data}
\label{app:train-data}

We post-train every method on \textbf{MixChain-Z-PRM12K}, a mathematical-reasoning corpus of roughly $12$K competition-style problems in the MATH/PRM800K \cite{prm800k} lineage. Each problem is paired with a checkable ground-truth final answer and one or more chain-of-thought (CoT) solutions; the ``mixed-chain'' construction supplies solutions of \emph{varying reasoning length} for the same problem, from terse derivations to fully elaborated step-by-step traces, which exposes the model to a range of solution granularities rather than a single canonical style. The problems span the standard competition-mathematics topics, arithmetic and algebra, number theory, counting and probability, and geometry, at a spread of difficulty levels.

Only two fields of each record are used by our method. The final answer feeds the rollout verifier of Eq.~\eqref{eq:ldopsd}, which checks a completion by comparing its extracted answer against this gold value; the verification is answer-level and never exposes the reference CoT to the model. The reference CoT solutions are used only by the baselines that require them: SFT imitates them token by token, and OPSD (full-solution) conditions its teacher on the complete reference solution as PI. dOPSD itself consumes neither the reference CoT (its privilege is trajectory-derived) nor, in the unverified variant of the ablation in Figure~\ref{fig:verify}, the final answer.

For all methods we format each problem with the same instruction-style prompt template used at evaluation, so that the training condition matches the inference condition the student is later scored under. Problems whose gold answer cannot be parsed into a checkable form are discarded. Prompts and completions are tokenized with each backbone's native tokenizer and truncated to the model's generation length; we use a completion budget of $256$ tokens, matching the diffusion decoding length at which the models are evaluated.

\subsection{Evaluation Benchmarks}
\label{app:eval-data}

We evaluate on four public benchmarks: two in-domain mathematical-reasoning sets and two out-of-distribution code-generation sets. None of them overlaps with the training corpus, and the code benchmarks in particular probe transfer to a domain never seen during post-training. Table~\ref{tab:datasets} lists their sizes, domains, and scoring metrics.

\begin{table}[t]
\centering
\setlength{\tabcolsep}{4pt}
\begin{tabular}{l l r l}
\toprule
Benchmark & Domain & \#Test & Metric \\
\midrule
\multicolumn{4}{l}{\textit{In-domain: mathematical reasoning}} \\
GSM8K & Grade-school math & $1{,}319$ & Acc@1 \\
MATH500 & Competition math & $500$ & Acc@1 \\
\midrule
\multicolumn{4}{l}{\textit{Out-of-distribution: code generation}} \\
HumanEval & Python synthesis & $164$ & pass@1 \\
MBPP & Python synthesis & $500$ & pass@1 \\
\bottomrule
\end{tabular}
\caption{Evaluation benchmarks. Math benchmarks are scored by final-answer accuracy (Acc@1); code benchmarks by functional correctness against the provided unit tests (pass@1). MBPP uses the sanitized test split.}
\label{tab:datasets}
\end{table}

\paragraph{GSM8K.} GSM8K consists of grade-school arithmetic word problems that require multi-step numerical reasoning. We evaluate on its $1{,}319$-problem test split and score a completion as correct when its extracted final answer exactly matches the gold value (Acc@1).

\paragraph{MATH500.} MATH500 is the $500$-problem subset of the MATH benchmark of competition mathematics, spanning seven subjects and five difficulty levels. Answers are checked for mathematical equivalence to the reference (Acc@1), so equivalent surface forms of the same value are counted as correct.

\paragraph{HumanEval.} HumanEval contains $164$ hand-written Python programming problems, each specified by a function signature, a natural-language docstring, and a set of hidden unit tests. A sample is correct if the generated function passes all of its tests; we report pass@1.

\paragraph{MBPP.} MBPP (Mostly Basic Python Problems) comprises short, entry-level Python tasks, each with a prompt and a small suite of unit tests. We evaluate on the sanitized test split of $500$ problems and report pass@1 under the same functional-correctness criterion as HumanEval.

For all four benchmarks we decode with the same diffusion sampler and generation length used during training and greedily read out the final answer, so the reported numbers reflect single-sample accuracy rather than any best-of-$n$ or self-consistency aggregation.

\section{Hyperparameters for Inference}
\label{app:infer}

Table~\ref{tab:infer-hparams} lists the full decoding configuration used for every evaluation, separately for each backbone and benchmark. All numbers in the paper use single-sample decoding, so the math benchmarks report Acc@1 and the code benchmarks pass@1; no best-of-$n$ or self-consistency is used.

Two invariants hold across all runs. First, the number of diffusion steps is set equal to the number of generated tokens on both backbones: letting the sampler commit more than one token per step degrades quality sharply, so we decode a single token per step. Second, the generation length is chosen per benchmark to comfortably cover the reference solutions while staying within each model's position limit, which is why code generation, with its longer completions, uses more tokens than the math benchmarks.

\begin{table*}[t]
\centering
\begin{tabular}{l cccc}
\toprule
Hyperparameter & GSM8K & MATH500 & HumanEval & MBPP \\
\midrule
\multicolumn{5}{l}{\textit{Dream-7B-Instruct}} \\
Max new tokens & $256$ & $512$ & $768$ & $1024$ \\
Diffusion steps & $256$ & $512$ & $768$ & $1024$ \\
Temperature & $0.1$ & $0.1$ & $0.1$ & $0.1$ \\
Top-$p$ & $0.9$ & $0.9$ & $0.9$ & $0.9$ \\
Top-$k$ & -- & -- & -- & -- \\
\midrule
\multicolumn{5}{l}{\textit{LLaDA-8B-Instruct}} \\
Max new tokens & $256$ & $512$ & $512$ & $256$ \\
Diffusion steps & $256$ & $512$ & $512$ & $256$ \\
Block length & $8$ & $64$ & $512$ & $256$ \\
Temperature & $0.0$ & $0.0$ & $0.0$ & $0.0$ \\
Top-$p$ & $0.9$ & $0.9$ & $0.9$ & $0.9$ \\
Top-$k$ & -- & -- & -- & -- \\
\bottomrule
\end{tabular}
\caption{Inference hyperparameters across base models and benchmarks. ``--'' indicates the hyperparameter is not applicable or not used. On both backbones the number of diffusion steps equals the number of generated tokens. For LLaDA, temperature is $0$ (greedy) and top-$p$ is informational only, as its Gumbel-argmax sampling ignores nucleus truncation.}
\label{tab:infer-hparams}
\end{table*}

\paragraph{Dream-7B-Instruct.} Dream decodes with confidence-based (entropy) remasking at a low temperature of $0.1$ and nucleus sampling with top-$p=0.9$, without top-$k$ truncation. For code generation on HumanEval we additionally enable entropy-threshold parallel decoding (threshold $0.5$) for efficiency; all other Dream runs use the plain entropy sampler.

\paragraph{LLaDA-8B-Instruct.} LLaDA decodes with its native low-confidence remasking at temperature $0$ (greedy). Because its Gumbel-argmax sampling is insensitive to nucleus truncation, the listed top-$p=0.9$ is informational only and does not affect the output. LLaDA additionally denoises in semi-autoregressive \emph{blocks}; we set the block length per benchmark following the model's recommended settings, from a fine block of $8$ on GSM8K to a single full-length block on HumanEval.

\section{Qualitative Examples}
\label{app:examples}

To make the baselines' failure modes concrete, we show every method's generation on three representative GSM8K problems (Dream-7B-Instruct). Each problem isolates a distinct error that the baselines share but dOPSD avoids. Throughout, highlighting marks the pivotal \hlok{correct} and \hlbad{erroneous} steps.

\subsection{Example 1: Problem 526}

The problem states a \emph{daily} consumption directly, so no hour-to-day conversion is needed. Nonetheless, every baseline hallucinates a spurious ``$\times 24$ hours'' step, and several also subtract the two bills in the wrong direction; the errors compound into wildly off answers. dOPSD is the only method that reads the units correctly and reaches the gold answer of $21$.

\begin{solbox}[Problem \hfill Gold answer: 21]{cneu}
Ada's daily electric consumption is 12 kilowatts per hour. She is planning to add a device that will consume 2 kilowatts per hour a day. If a kilowatt per hour is \$1.50, how much is the difference between Ada's weekly electric bill before and after she adds the new device?
\end{solbox}

\begin{solbox}[dOPSD \hfill Predicted: 21 ~$\checkmark$]{cok}
Ada's daily electric consumption is 12 kilowatts per hour $\times$ \$1.50 = \hlok{\$18 per day}. Her weekly bill is \$18 $\times$ 7 days = \$126. Adding the new device, her daily consumption is $12 + 2 = 14$ kilowatts per hour, so her weekly bill is 14 $\times$ \$1.50 $\times$ 7 days = \$147. The difference is \$147 $-$ \$126 = \hlok{\$21}. \textbf{The answer: 21.}
\end{solbox}

\begin{solbox}[Base model \hfill Predicted: 504 ~$\times$]{cbad}
Ada's daily electric consumption is 12 kilowatts per hour \hlbad{$\times$ 24 hours} $= 288$ kilowatts per day. Her daily bill is 288 $\times$ \$1.50 = \$432, so her weekly bill is \$432 $\times$ 7 = \$3024. The new device consumes 2 \hlbad{$\times$ 24} $= 48$ kilowatts per day, giving $288 + 48 = 336$ kilowatts per day, \$504 per day, \$3528 per week. The difference is \$3528 $-$ \$3024 = \textbf{\$504.}
\end{solbox}

\begin{solbox}[SFT \hfill Predicted: 2520 ~$\times$]{cbad}
Ada's daily consumption is 12 \hlbad{$\times$ 24 hours} $= 288$ kilowatts per day, \$432 per day, \$3024 per week. The new device consumes 2 \hlbad{$\times$ 24} $= 48$ kilowatts per day, \$72 per day, \$504 per week. The difference is \hlbad{\$3024 $-$ \$504} = \$2520. \textbf{The answer: 2520.}
\end{solbox}

\begin{solbox}[GRPO \hfill Predicted: 2877 ~$\times$]{cbad}
Ada's daily consumption is 12 \hlbad{$\times$ 24 hours} $= 288$ kilowatts per day, \$432 per day, \$3024 per week. Then her daily consumption is $12 + 2 = 14$ kilowatts per hour, \$21 per day, \$147 per week. The difference is \hlbad{\$3024 $-$ \$147} = \$2877. \textbf{The answer: 2877.}
\end{solbox}

\begin{solbox}[OPSD (answer-only) \hfill Predicted: 504 ~$\times$]{cbad}
Ada's daily consumption is 12 \hlbad{$\times$ 24 hours} $= 288$ kilowatts per day, \$432 per day, \$3024 per week. The new device consumes 2 \hlbad{$\times$ 24} $= 48$ kilowatts per day, giving $336$ kilowatts per day, \$504 per day, \$3528 per week. The difference is \$3528 $-$ \$3024 = \textbf{\$504.} (Identical hallucinated conversion as the base model.)
\end{solbox}

\begin{solbox}[OPSD (full-solution) \hfill Predicted: none ~$\times$]{cbad}
Ada's daily consumption is 12 \hlbad{$\times$ 24} $= 288$ kilowatts per day. After adding the device, $288 + 2 = 290$ kilowatts per day. The weekly bills are $288 \times 1.50 = \$432$ and $290 \times 1.50 = \$435$, so the difference is \$435 $-$ \$432 = \$\dots{} \emph{(the verbose generation exceeds the budget and terminates without a final answer).}
\end{solbox}

Only dOPSD avoids the spurious hourly conversion: its trajectory-derived teacher keeps the student on unit-consistent reasoning, whereas every supervised, reinforcement-learning, and PI-conditioned baseline inherits or amplifies the $\times 24$ error, and the richest-privilege variant (full-solution) additionally rambles into an unfinished, malformed answer.

\subsection{Example 2: Problem 589}

Here one option makes Martin \emph{lose} weight and the other makes him \emph{gain} weight, so the two outcomes lie on opposite sides of his starting weight. The correct gap therefore treats the Cheerios option as $-6.25$ and the donut option as $+8.75$, giving $8.75-(-6.25)=15$. Every baseline instead subtracts the two magnitudes, $8.75-6.25=2.5$, silently dropping the sign; dOPSD alone recognizes that a loss and a gain add up and returns the gold answer $15$.

\begin{solbox}[Problem \hfill Gold answer: 15]{cneu}
If Martin eats Cheerios every day for breakfast, he'll lose 1.25 pounds/week. If he eats donuts every day for breakfast, he'll gain 1.75 pounds/week. What will be the difference in his weight at the end of 5 weeks between the two breakfast options?
\end{solbox}

\begin{solbox}[dOPSD \hfill Predicted: 15 ~$\checkmark$]{cok}
If Martin eats Cheerios, he'll lose 1.25 pounds/week $\times$ 5 weeks = 6.25 pounds. If he eats donuts, he'll gain 1.75 pounds/week $\times$ 5 weeks = 8.75 pounds. Because one option is a loss and the other a gain, the difference at the end of 5 weeks is \hlok{15 pounds}. \textbf{The answer is 15.}
\end{solbox}

\begin{solbox}[Base model \hfill Predicted: 2.5 ~$\times$]{cbad}
If Martin eats Cheerios, he'll lose 1.25 pounds/week $\times$ 5 weeks = 6.25 pounds. If he eats donuts, he'll gain 1.75 pounds/week $\times$ 5 weeks = 8.75 pounds. The difference is \hlbad{8.75 $-$ 6.25 = 2.5 pounds}. \textbf{The answer is 2.5.}
\end{solbox}

\begin{solbox}[SFT \hfill Predicted: 2.5 ~$\times$]{cbad}
Over 5 weeks the Cheerios option loses $1.25 \times 5 = 6.25$ pounds and the donut option gains $1.75 \times 5 = 8.75$ pounds. The difference between the two options is \hlbad{8.75 $-$ 6.25 = 2.5 pounds}. \textbf{The answer: 2.5.}
\end{solbox}

\begin{solbox}[GRPO \hfill Predicted: 2.5 ~$\times$]{cbad}
After 5 weeks Martin has lost $1.25 \times 5 = 6.25$ pounds on Cheerios and gained $1.75 \times 5 = 8.75$ pounds on donuts. The difference at the end of 5 weeks is \hlbad{8.75 $-$ 6.25 = 2.5 pounds}. \textbf{The answer: 2.5.}
\end{solbox}

\begin{solbox}[OPSD (answer-only) \hfill Predicted: 2.5 ~$\times$]{cbad}
Over 5 weeks the Cheerios option loses $6.25$ pounds and the donut option gains $8.75$ pounds. The difference is \hlbad{8.75 $-$ 6.25 = 2.5 pounds}; the model then restates the same value verbatim. \textbf{The answer is: 2.5.}
\end{solbox}

\begin{solbox}[OPSD (full-solution) \hfill Predicted: none ~$\times$]{cbad}
\emph{Empty response: the model generated no text at all (raw output \texttt{['\,']}).}
\end{solbox}

\subsection{Example 3: Problem 780}

The question asks for the \emph{difference} between two scores, which is a magnitude. Every method computes the signed gap correctly ($80+15-120=-25$, i.e.\ Ava scores $25$ below Emma), but only dOPSD reports it as the positive magnitude $25$; the baselines box the raw signed value $-25$, discard it while second-guessing the unknown absolute scores, or collapse to $0$.

\begin{solbox}[Problem \hfill Gold answer: 25]{cneu}
Ava and Emma want to know who is better at a new video game. They each play one level; the higher score wins. They receive 10 points per enemy jumped on, 5 points per berry collected, and 30 points for every second left on the timer. If Ava jumps on 8 more enemies than Emma and collects 3 more berries, but finishes the level 4 seconds slower, what is the difference between their two scores?
\end{solbox}

\begin{solbox}[dOPSD \hfill Predicted: 25 ~$\checkmark$]{cok}
Ava gains $8\times10=80$ (enemies) and $3\times5=15$ (berries) but loses $4\times30=120$ (timer), so the signed gap is $80+15-120=-25$. Since it is negative, Ava's score is lower than Emma's, and the difference between their scores is \hlok{$25$ points}. $\boxed{25}$
\end{solbox}

\begin{solbox}[Base model \hfill Predicted: -25 ~$\times$]{cbad}
Ava gains $80$ (enemies) and $15$ (berries) but loses $120$ (timer): $80+15-120=-25$. The difference between their scores is reported as the signed value \hlbad{$\boxed{-25}$}.
\end{solbox}

\begin{solbox}[SFT \hfill Predicted: -25 ~$\times$]{cbad}
Computes Ava's net as $80+15-120=-25$, then reasons that Emma's absolute score is unknown, so the difference ``cannot be found,'' expressing it as $-25$ less than Emma's score. \hlbad{The answer is: $-25$.}
\end{solbox}

\begin{solbox}[GRPO \hfill Predicted: 0 ~$\times$]{cbad}
Computes Ava's net as $80+15-120=-25$, introduces an unknown Emma score $x$, writes the difference as $x-(-25)=x+25$, then, unable to determine $x$, collapses to \hlbad{$\boxed{0}$}.
\end{solbox}

\begin{solbox}[OPSD (answer-only) \hfill Predicted: -25 ~$\times$]{cbad}
Same computation, $80+15-120=-25$, reported as the signed value \hlbad{$\boxed{-25}$}.
\end{solbox}

\begin{solbox}[OPSD (full-solution) \hfill Predicted: none ~$\times$]{cbad}
\emph{Empty response: the model generated no text at all (raw output \texttt{['\,']}).}
\end{solbox}

Across all three examples the pattern is the same: the baselines make a single, confident slip, a hallucinated unit conversion, a dropped sign, or a mis-reported signed difference, that the final answer inherits, while the full-solution teacher degrades into malformed or empty generations. dOPSD, drawing its teacher signal from the model's own decoding trajectory rather than an external solution, is the only method that keeps the reasoning faithful and lands on the gold answer in every case.

\end{document}